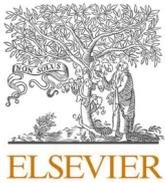
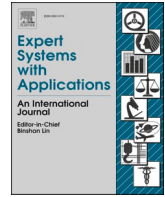

# Deep learning in computed tomography pulmonary angiography imaging: A dual-pronged approach for pulmonary embolism detection

Fabiha Bushra [a], Muhammad E.H. Chowdhury [b,*], Rusab Sarmun [a], Saidul Kabir [a], Menatalla Said [c], Sohaib Bassam Zoghoul [d], Adam Mushtak [d], Israa Al-Hashimi [d], Abdulrahman Alqahtani [e,f], Anwarul Hasan [g]

[a] *Department of Electrical and Electronic Engineering, University of Dhaka, Dhaka 1000, Bangladesh*
[b] *Department of Electrical Engineering, Qatar University, Doha 2713, Qatar*
[c] *Department of Basic Medical Sciences, College of Medicine, Qatar University, Doha 2713, Qatar*
[d] *Department of Radiology, Hamad Medical Corporation, Doha, Qatar*
[e] *Department of Biomedical Technology, College of Applied Medical Sciences in Al-Kharj, Prince Sattam Bin Abdulaziz University, Al-Kharj 11942, Saudi Arabia*
[f] *Department of Medical Equipment Technology, College of Applied, Medical Science, Majmaah University, Majmaah City 11952, Saudi Arabia*
[g] *Department of Industrial and Mechanical Engineering, Qatar University, Doha 2713, Qatar*

ABSTRACT

The increasing reliance on Computed Tomography Pulmonary Angiography (CTPA) for Pulmonary Embolism (PE) diagnosis presents challenges and a pressing need for improved diagnostic solutions. The primary objective of this study is to leverage deep learning techniques to enhance the Computer Assisted Diagnosis (CAD) of PE. With this aim, we propose a classifier-guided detection approach that effectively leverages the classifier's probabilistic inference to direct the detection predictions, marking a novel contribution in the domain of automated PE diagnosis. Our classification system includes an Attention-Guided Convolutional Neural Network (AG-CNN) that uses local context by employing an attention mechanism. This approach emulates a human expert's attention by looking at both global appearances and local lesion regions before making a decision. The classifier demonstrates robust performance on the FUMPE dataset, achieving an AUROC of 0.927, sensitivity of 0.862, specificity of 0.879, and an F1-score of 0.805 with the Inception-v3 backbone architecture. Moreover, AG-CNN outperforms the baseline DenseNet-121 model, achieving an 8.1% AUROC gain. While previous research has mostly focused on finding PE in the main arteries, our use of cutting-edge object detection models and ensembling techniques greatly improves the accuracy of detecting small embolisms in the peripheral arteries. Finally, our proposed classifier-guided detection approach further refines the detection metrics, contributing new state-of-the-art to the community: $mAP_{50}$, sensitivity, and F1-score of 0.846, 0.901, and 0.779, respectively, outperforming the former benchmark with a significant 3.7% improvement in $mAP_{50}$. Our research aims to elevate PE patient care by integrating AI solutions into clinical workflows, highlighting the potential of human-AI collaboration in medical diagnostics.

## 1. Introduction

Pulmonary embolism (PE) is a life-threatening medical condition characterized by the obstruction of one or more pulmonary arteries due to the presence of emboli, most commonly originating from deep vein thrombosis (DVT) (Ran et al., 2022). This cardiovascular disorder poses a significant global health burden and is associated with severe complications, including respiratory distress, right ventricular dysfunction, and even sudden death (Longo et al., 2011). It also poses a significant burden on healthcare systems worldwide, with an annual mortality rate of 100,000 to 300,000 deaths in the United States alone (Abd El Halim et al., 2014). Unfortunately, PE is widely recognized for its propensity to be underdiagnosed, which can significantly impact patient care by leading to delays in treatment and exacerbating clinical outcomes (Raja et al., 2015). While quick and precise diagnosis is crucial, many healthcare systems and radiology providers find it challenging to maintain, especially as its demand has grown; for instance, the use of Computed Tomography Pulmonary Angiography (CTPA) in emergency situations has surged 27 times in the last 20 years (Chandra et al., 2013; Prologo et al., 2004).

Recent studies have demonstrated that machine learning and deep learning-based technologies have already exhibited considerable potential in the Computer Assisted Diagnosis (CAD) of medical images (Chowdhury et al., 2020; Rahman et al., 2020; Asman et al., 2015). While traditional machine learning methods have played a significant role in the image processing domain (Luo et al., 2019; Zheng et al., 2019), deep learning, a subset of machine learning, has gained particular attention due to its ability to automatically learn and extract highly relevant features directly from images, eliminating the need for manual feature extraction and leveraging the advantages of deep neural networks in challenging tasks such as complex medical image analysis (Tahir et al., 2021). Although there are evident clinical and engineering benefits in using deep learning for automated disease identification on CTA scans (Qiblawey et al., 2021; Khan et al., 2023), considerable challenges persist in the case of CTPA analyses. One of the primary challenges in detecting PE through CTA DICOM images is differentiating filling defects caused by PE from other potential causes of chest pain, such as pneumonia or pulmonary edema (Moore et al., 2018). Traditional CT scans take longer and can produce respiratory motion artifacts, reducing the image quality and its diagnostic usefulness for PE identification (Rodríguez-Romero and Castro-Tejero, 2014). Furthermore, the detection of small embolisms in the peripheral arteries is challenging since they frequently resemble small blood vessels, making them difficult to differentiate (Long et al., 2021). Lastly, achieving consistency across different institutions, especially with diverse CT scanner models and reconstruction techniques, poses a challenge for automated diagnostic generalization.

Despite the inherent challenges, leveraging deep learning for an automated PE diagnosis through CTPA holds immense potential in terms of,

- **Prevalence & Severity.** PE is a prevalent and potentially fatal condition, making swift and accurate diagnosis paramount for medical professionals, patients, insurance companies, and regulatory agencies.
- **CTPA's Dominance.** As the gold standard, CTPA remains the most trusted and widely used imaging modality for PE diagnosis (Moore et al., 2018), emphasizing its pivotal role in medical diagnostics.
- **Self-sufficient Diagnosis.** A conclusive diagnosis of PE can be determined through CT imaging, eliminating the need for additional diagnostic steps or pathological validation which aligns well with a supervised learning approach.

Deep learning models become increasingly prominent in medical image analysis promoting a collaborative environment between humans and machines (Guan et al., 2022). This collaboration not only streamlines routine tasks for overworked medical professionals but also intelligently prioritizes radiological analyses for quicker diagnosis of critical pathologies. Furthermore, recent studies highlight the significant role of deep learning approaches in applications as disparate as gait pattern recognition to cardiac electrodynamics modeling (Chen et al., 2021; Tian et al., 2022; Xu et al., 2022; Xie and Yao, 2022), showcasing their adaptability and effectiveness. Such advancements motivate the development of innovative deep learning algorithms for emergent scenarios, including PE detection in CTPA studies.

Prior research on machine learning (ML) for PE detection had focused on using CAD software to identify PE candidates from CT images, even though CAD software had low sensitivity and specificity requiring manual verification by radiologists (Lee et al., 2011). Early efforts on automated PE diagnosis also included using electrocardiogram (ECG) signals to predict PE likelihood from clinical data; however, ECG signals had low diagnostic accuracy and could be affected by other cardiac conditions (Somani et al., 2022). Other works also included using deep neural networks (DNNs) to detect PE from CT images which required large amounts of data and computational resources, and could suffer from overfitting and generalization issues (Huhtanen et al., 2022). In contrast, recent studies have shifted focus to diagnosing PE by using CNNs on CTPA images; for instance, Tajbakhsh et al. proposed 3D CNNs on CT scans to detect PE in 2019, utilizing a comprehensive set of pre-processed features from segmentation and vessel alignment as input to their CNN framework (Tajbakhsh et al., 2015). Since then, researchers have conducted comprehensive analyses on the classification, detection, and segmentation of CTPA images using deep learning techniques to identify PE.

Long et al. introduced a P-Mask R-CNN model for PE detection using probability-based approaches (Long et al., 2021). The model uses upsampling to enrich the anchor's detail, emphasizing small object detection. By choosing anchors from probable PE areas on the feature map, it minimizes redundant anchors and attains an Average Precision (AP) of 80.9 % at a 0.5 IoU threshold on the FUMPE dataset (Masoudi et al., 2018). While the study emphasizes small embolism detection, it neglects the vital aspect of integrating arterial context to distinguish these embolisms from visually similar irregularities, such as dark veins and lymph tissues which can exhibit similarities to PE regions, especially when lacking contrast enhancement (Teigen et al., 1995). Another research developed fusion models that combined CT scans and Electronic Medical Records (EMRs) for automated PE classification (Huang et al., 2020). They used a 77-layer 3D CNN, PENet (Huang et al., 2020), for CT data and applied feature engineering for EMRs. Three fusion strategies—early, late, and joint—were explored to integrate both data sources, achieving an Area Under the Receiver Operating Characteristic (AUROC) of 0.962. However, their study predominantly approaches the problem as a classification task, which limits its capacity to precisely locate emboli. Recently, one study proposed a deep neural network model to classify PE from CTPAs using a small dataset of 600 CTPAs with weakly labeled data (Huhtanen et al., 2022). The authors employed a 2D CNN based on InceptionResNetV2 to analyze CTPA slices and a Long Short-Term Memory Network (LSTM) to process these in sequence. Nonetheless, the reliance on weakly labeled data in a medical context, where high sensitivity is imperative, leads to suboptimal detection of PE, as this approach does not capture the full complexity and variability of the condition. Ma et al. presented a two-phase multitask deep learning method for classifying PE from 3D CTPA images (Ma et al., 2022). Initially, a fine-tuned 3D ResNet-18 model processes these images, extracting features from the 3D windows which then move to the second phase, where a Temporal Convolutional Network (TCN) handles them to predict study-level labels related to PE achieving an AUROC of 0.926. Utilizing a cloud-based PE detection algorithm, Weikert et al. developed a region-based CNN that trained on 28,000 CTPAs from nine medical centers of which 43.4 % contained PE slices (Weikert et al., 2020). The two-stage algorithm first employed a 3D CNN based on the Resnet architecture, to generate a segmentation map. The second stage classified regions as positive or negative for PE. A novel CNN architecture, PE-DeepNet, was presented in the literature (Lynch and Suriya, 2022) for classifying CTPA images using the RSNA PE dataset (Colak et al., 2021). PE-DeepNet features two 2D convolutional layers with Rectified Linear Unit (ReLU) activation function, a max-pooling layer, a unique PE-DeepNet block, and a sigmoid classifier. In another study, a CNN model was employed in which Maximum intensity projection (MIP) input images were processed through DenseNet-121, and the averaged feature vectors were subsequently passed to a three-layer classification network (Vainio et al., 2023). The network predicted the presence of Chronic PE for each lung, using the maximum of the two predictions as the final disease indicator. However, the majority of these studies focus solely on the global CTPA image for PE diagnosis, often lacking the necessary attention to specific lung regions required for precise diagnoses. Moreover, these studies primarily concentrate on detecting embolisms in the main artery while disregarding the smaller embolisms present in peripheral arteries. As a consequence, these prior approaches lack the crucial sensitivity required in the medical diagnostic context, where failing to identify positive cases can have significant repercussions.

In this study, we aim to bridge critical research gaps in the field of PE detection using deep learning techniques. With this objective, we propose a classifier-guided detection approach that effectively leverages the classifier's probabilistic inference to refine the detection predictions. Our integrated dual-pronged approach combining classification and detection marks a novel contribution in the domain of automated PE diagnosis, distinct from existing approaches in the literature. Furthering this contribution, our classification framework addresses the need for a comprehensive methodology that employs an attention mechanism to extract contextual information from local lesion regions, a sharp contrast with existing research efforts that rely solely on global appearances for predictions. In addition, our detection framework deals with the key limitation of accurately localizing small embolisms in the peripheral arteries. While prior studies have primarily focused on PE detection in main arteries, our utilization of state-of-the-art object detection models and ensembling techniques significantly enhances detection accuracy for small embolisms existing in the peripheral arteries. This approach takes a vital research gap into account, as accurately detecting such embolisms is crucial for achieving a comprehensive diagnosis across a wide range of cases. Furthermore, an innovative strategy that our study utilizes is the incorporation of essential arterial context data for training the detection models. As previous studies have included embolisms within tightly defined bounding boxes, they often lack the broader context of surrounding arteries and blood vessels. Our strategy addresses this limitation by providing the necessary arterial information to the model, thereby enhancing its ability to make more informed and accurate detections. By addressing the prevailing challenges of automated PE diagnosis, our study aims to elevate the standard of care for PE patients and emphasizes the advantages of integrating AI solutions into clinical workflows.

The major contributions of our study can be summarized as,

- We propose a deep learning-based classification approach that leverages local context by utilizing an attention mechanism for improved diagnosis of PE. This framework emulates the attention of a human expert by considering both global appearances and local lesion regions before forming a conclusive decision.
- We demonstrate major improvements over baseline models by incorporating the attention method on the classification framework; improving AUROC by 8.1 % on a publicly available CTA dataset of PE.
- We employ state-of-the-art object detection models to localize embolisms in both the main and peripheral arteries. The implementation of ensembling techniques significantly enhances the detection accuracy, especially for detecting the small embolisms residing in the peripheral arteries—a crucial aspect for the effective diagnosis of a diverse range of embolisms.
- To counter the false positives associated with the high sensitivity of the detection models, we adopt a heuristic strategy, intertwining classifier outcomes with detection results for a more precise PE diagnosis.

In this paper, the content is divided into five sections. The models used for PE classification and detection are discussed in the second section. The experimental methodology of the research is discussed in detail in the third section. The fourth part presents the results of the study and a comprehensive analysis of the performance of the models, and the fifth section concludes the paper.

## 2. Models for image classification and detection

In this section, we present an overview of the key models utilized for image classification and detection in this study, highlighting their unique architectures and essential functionalities.

### 2.1. AG-CNN

The architecture, named Attention Guided Convolutional Neural Network (AG-CNN), is a classification framework that employs an attention mechanism to create a binary mask on the feature maps, focusing on regions of the image that hold the most diagnostic significance (Guan et al., 2020). The model architecture consists of three main branches: the global branch, the local branch, and the fusion branch. The global branch processes the whole input image and predicts the presence of pathologies. The output of this branch informs the underlying information of the global image as a whole. The local branch focuses on the lesion area, which is expected to alleviate the drawbacks of only using the global image. The local branch uses the same network architecture as the global branch, but their weights are not shared due to their different roles in the model. The input to the local branch is a cropped image from the global image representing an attractive region inferred by an attention-guided mask inference process. The binary mask is constructed by performing Otsu's thresholding operations on the feature maps for employing the attention process (Liu and Yu, 2009). If the value of a certain spatial position in the heatmap exceeds a threshold, a value of 1 is assigned to the corresponding position in the mask. This signifies that attention or focus should be directed to this region. Ultimately, the fusion branch integrates the outputs from the global and local branches by concatenating their last pooling layers (Pool5) and fine-tuning them, resulting in the final prediction. The architecture for AG-CNN with Inception-v3 as the backbone is shown in Fig. 1 (Szegedy et al., 2016).

### 2.2. Yolov8

YOLOv8 is the most recent addition to the YOLO (You Only Look Once) series of object detection models developed by Ultralytics (Jocher et al., 2023). The model adopts an anchor-free approach to directly predict object centers unlike the previous versions of YOLO that relied on anchor boxes (Solawetz and Francesco, 2023). This eliminates the need for anchor box design, which was a challenging aspect in earlier YOLO versions, as anchor boxes might not represent the distribution of custom datasets accurately. By using an anchor-free method, YOLOv8 reduces the number of box predictions, leading to faster Non-Maximum Suppression (NMS) during post-processing. The backbone of the architecture is responsible for extracting relevant features from the input image, and it includes changes such as using a 3x3 convolution instead of the 6x6 convolution in the stem and altering the main building block. The Spatial Pyramid Pooling Fusion (SPPF) layer takes the feature maps generated by the backbone and performs spatial pyramid pooling on them. The Context-to-Filter (C2F) module, a novel addition, replaces the traditional YOLO neck architecture. In YOLOv8, the head is engineered to operate in a decoupled manner. This design permits separate processing of objectness, classification, and regression tasks, thereby enabling each branch to concentrate exclusively on its specific task. To address the class imbalance in the training data, YOLOv8 uses the Distribution Focal Loss (DFL) for the localization loss. DFL discretizes the predicted geometrical features into bins and gives more weight to hard-to-classify examples. YOLOv8 introduces the use of a larger Stride parameter for smaller input image sizes, striking a balance between detecting smaller objects and maintaining a reasonable output resolution and receptive field size.

### 2.3. Faster R-CNN

Faster R-CNN is an object detection model that consists of two main components: a region proposal network (RPN) and a region-based CNN (RCNN). The RPN generates a set of candidate regions of interest (ROIs) that may contain objects, while the RCNN classifies the ROIs and refines their bounding boxes. Both components share the same backbone network that extracts features from the input image. Faster R-CNN is an improvement over Fast R-CNN, which used a separate method for region proposal generation that was slow and costly. The RPN is a fully convolutional network that predicts object bounds and objectness scores at each position of the feature map. It uses a set of predefined anchor boxes with different scales and aspect ratios to generate ROIs. The RCNN is a network that takes the ROIs generated by the RPN as input and performs two tasks: classification and bounding box regression. The RCNN predicts four coordinates for each ROI, representing the refined bounding box location relative to the initial ROI. By sharing the convolutional features between the RPN and the RCNN, Faster R-CNN achieves faster and more accurate object detection than Fast R-CNN (Ren et al., 2015).

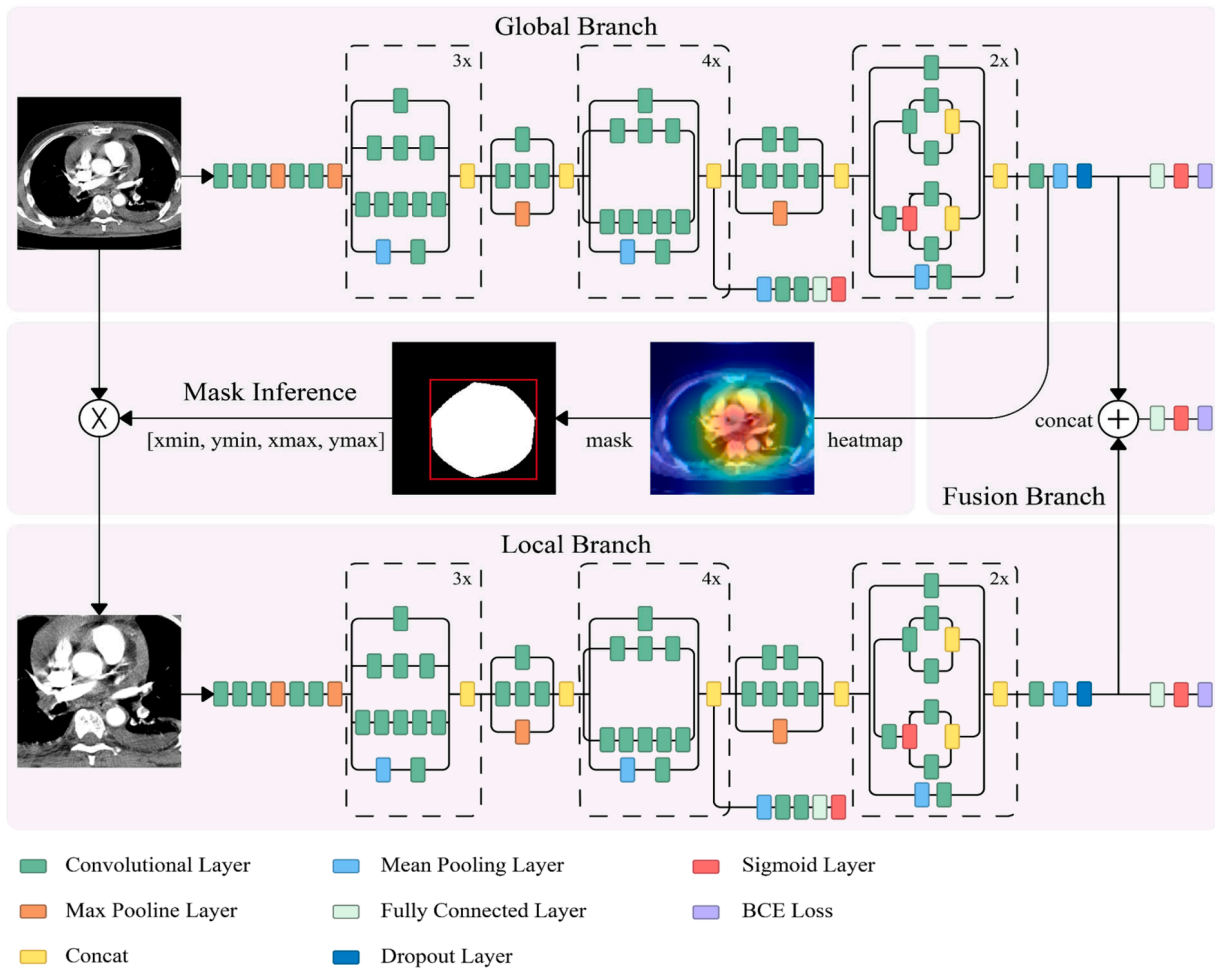

Fig. 1. AG-CNN architecture with Inception-v3 as the backbone.

### 2.4. EfficientDet

EfficientDet is a scalable and efficient object detection model that consists of three main components: an EfficientNet backbone, a BiFPN feature network, and a shared class/box prediction network. The EfficientNet backbone is a CNN that is pre-trained on ImageNet and provides multi-scale feature maps for the feature network. The BiFPN feature network is a weighted bi-directional feature pyramid network that fuses features from different levels and directions to produce a list of fused features for the prediction network. The shared class/box prediction network uses fused features to predict the class and location of each object in the image. EfficientDet also employs a compound scaling method that uniformly scales the resolution, depth, and width of all the components at the same time, allowing it to adapt to different resource constraints (Tan et al., 2020; Mingxing Tan, 2020; EfficientDet For PyTorch, 2023).

## 3. Experimental methodology

This section discusses the dataset used for this study (Masoudi et al., 2018) and the experimental steps involved in implementing our proposed framework.

### 3.1. Dataset description

The dataset used in this research work is a publicly accessible dataset called Ferdowsi University of Mashhad's PE (FUMPE) dataset specifically designed for CAD of PE using CTA images (Masoudi et al., 2018). In total, the dataset comprises 8792 slices acquired from 35 patients, out of which 2304 slices include lesion areas, and these slices contain a combined count of 3438 PE regions of interest (PE-ROIs). A significant quantity of PE regions, with approximately 67 % of the total PE-ROIs are situated in the lung's peripheral arteries. Notably, two of the patients had no PE clots in both the main and peripheral arteries. An overview of the dataset is given in Table 1 including patients' demographic data and site of PE-ROIs. Notably, the ages of two patients were not included in the dataset. Fig. 2 presents the distribution of PE-ROIs by size, with those having a square root area of only ≤ 30 pixels accounting for 83.33 % of the dataset.

### 3.2. Pre-Processing steps

**Window Width (WW) and Window Level (WL) Adjustment** In medical imaging, windowing is a technique used to enhance the visualization of specific tissues or structures within an image by adjusting the display range of Hounsfield Units (HU) captured in the Digital Imaging and Communications in Medicine (DICOM) format (Mandell et al., 2017). DICOM images, such as those obtained from CTA scans, are inherently represented in a wide range of HU values, each corresponding to different tissue densities (M. A, B. Y, and K. J., 2017). The windowing process allows the extraction of relevant information from these images by defining a specific window level (WL) and window width (WW) (Lee et al., 2018). For this study, a soft-tissue window was employed with a window level of 40 HU and a window width of 400 HU for the visual analysis of PE and other relevant lung pathologies (Lee et al., 2018).

**Table 1**
Overview of the FUMPE Dataset.

|  | Frequency | Percent |
| --- | --- | --- |
| **Gender:** |  |  |
| Male | 17 | 48.57 |
| Female | 18 | 51.43 |
| **Age:** |  |  |
| 18–40 | 10 | 21.62 |
| 41–65 | 7 | 32.43 |
| 65+ | 16 | 48.65 |
| **Site of PE-ROI:** |  |  |
| Main Artery | 1134 | 32.98 |
| Peripheral Artery | 2304 | 67.02 |

**Data Annotation** For the classification task, each image is represented by a two-dimensional one-hot encoded label vector, $L = [l_1, l_2]$, where each label $l_c \in \{0, 1\}$ and $C = 2$. The two labels correspond to the "Pulmonary Embolism" and "Non-Pulmonary Embolism" classes. If the corresponding mask of a CTA image contains a binary image that is non-blank, the CTA image is classified as "Pulmonary Embolism". Conversely, if the mask associated with the CTA image is a blank binary image, the CTA image is categorized as "Non-Pulmonary Embolism". In the process of dataset preparation for the detection task, the LabelImg tool was utilized for annotation (HumanSignal., (14–8-2023).). Traditional practices in object detection tasks involve creating bounding box annotations that tightly encompass the boundaries of the object of interest which in this case is an embolus. However, in this study, an alternative approach was adopted. The bounding box annotations were deliberately extended to incorporate additional pixels surrounding each embolus. This strategy was employed to offer the model supplemental arterial context through the inclusion of the surrounding arterial structures in the vicinity of the embolus. By enlarging the bounding box to encompass these additional pixels, the model is supplied with not only the visual information of the embolus but also its immediate arterial environment. Fig. 3(c) illustrates samples of these expanded bounding box annotations.

### 3.3. Experimental details

In this study, the patient-wise data splitting approach was adopted to prevent bias and data leakage, instead of random splitting of images which can lead to inflated performance metrics. By using patient-wise data splitting, each patient's data is exclusively in either the training or testing set, enabling a robust evaluation of the model's generalization ability to unseen patients (Bussola et al., 2021). For the classification task, the dataset comprising image slices from 35 patients, was divided into training and test sets in a ratio of 80:20, respectively. As a result, the training set for classifying PE contained 6892 images from 28 patients (80 %) and the test set contained 1900 images from 7 patients (20 %). Similarly, for the detection task, the dataset was divided into training, validation, and test sets with a patient-wise split ratio of 70:10:20, respectively leading to the training set including 6219 images from 25 patients (70 %), the validation set including 673 images from 3 patients (10 %), and the test set including 1900 images from 7 patients (20 %). To balance the training dataset between the PE class and the Non-PE class, data augmentation techniques encompassing random rotation, horizontal flip, random affine translations, and scaling transformations were employed. The experimental hardware setup, employed via Google Colab, comprised a single NVIDIA Tesla T4 with 15 GB GPU memory, a 2-core Intel Xeon CPU @ 2.00Ghz and 26 GB of system memory. All experiments were conducted using Python version 3.9.16, and Pytorch version 1.13.

### 3.4. Proposed method

In this study, we propose a comprehensive methodology for the diagnosis of PE that leverages a dual-pronged approach: classification and detection. The classification step applies an Attention-Guided Convolutional Neural Network (AG-CNN) to analyze the lung CTA images. This classification stage assists in evaluating whether the given image has features indicative of PE. To pinpoint the exact location of the potential embolism within the lungs, we employ the detection step. In this phase, state-of-the-art detection models are utilized to identify specific regions in the medical images that could be indicative of PE. Finally, we adopt a heuristic strategy by intertwining classifier outputs with detection results which sets a new benchmark in the field. If the classification stage indicates a high likelihood of PE, all detected embolisms are considered. However, if the likelihood is low, only the detections with a high level of confidence are kept. This comprehensive approach, integrating both classification and detection, allows for a robust and more accurate diagnosis of PE.

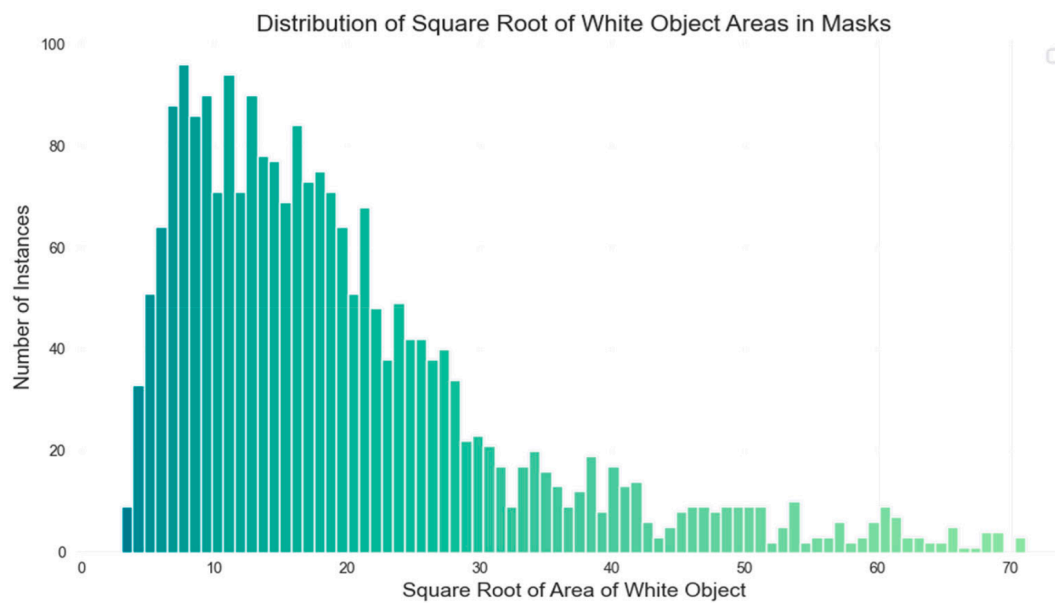

**Fig. 2.** Distribution of PE-ROI sizes present in the dataset. The majority of embolisms are too small for detection.

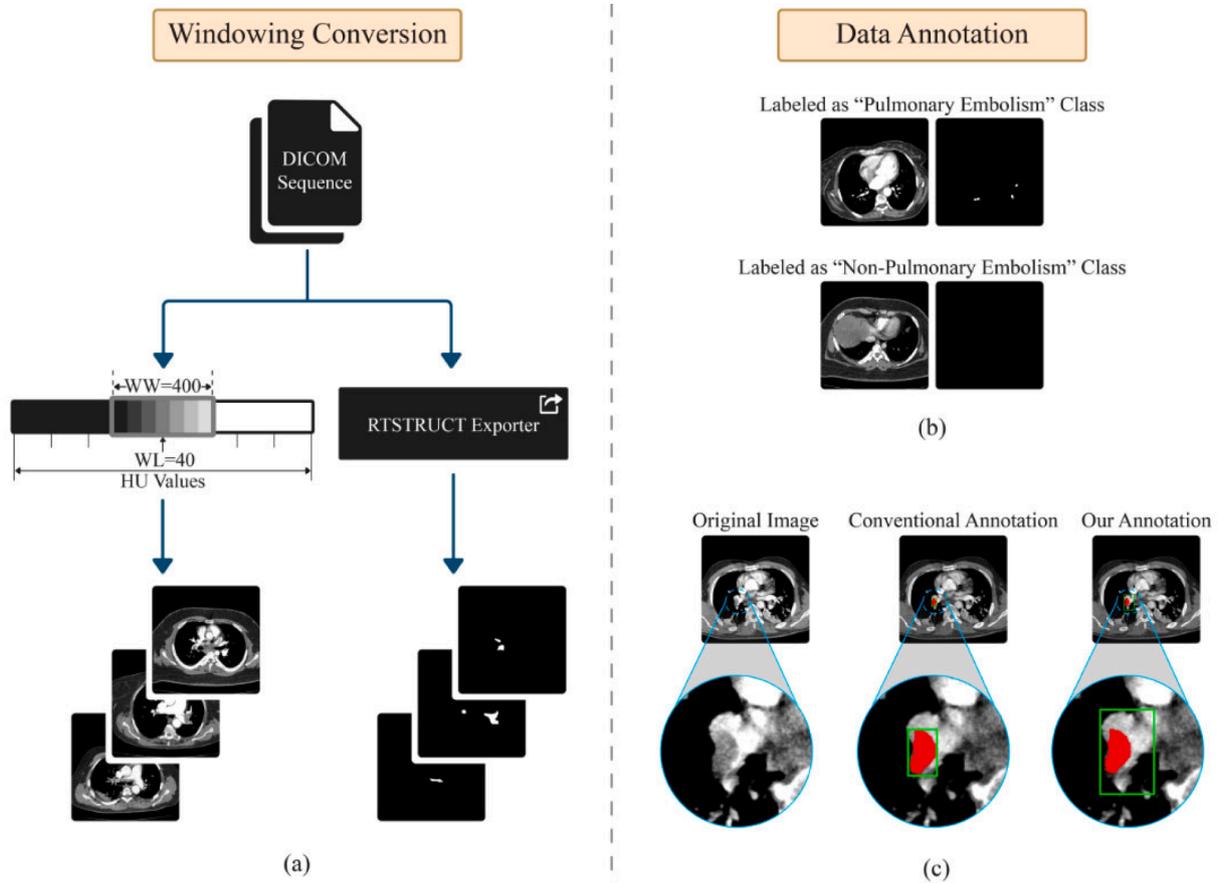

**Fig. 3.** Representation of Data Pre-processing Steps. (a) Generation of image slices using the "Soft Tissue Window" configuration with corresponding masks. (b) Image labeling based on the pixel values of the corresponding masks for the classification task; images with segmented masks are labeled as "Pulmonary Embolism" while images with blank masks are labeled as "Non-Pulmonary Embolism". (c) Image annotations for the detection task; bounding box annotations are extended around the emboli, differing from the conventional annotation approach that closely wraps the target object. The lesion area has been marked red for the visualization of the embolus.

**Classification Framework** For the classification task, we adopt an approach similar to that proposed by Guan et al. (2023) in their thorax disease classification study (Guan et al., 2020). In their work, they employed ResNet-50 as the primary backbone model for their AG-CNN framework, while also demonstrating the results achieved with DenseNet-121 used as another backbone model. However, we extend the scope by incorporating a broader range of classification models as backbone architectures, including ResNet-152, DenseNet201, Inception-v3, and MobileNetV3-Large (Szegedy et al., 2016; He et al., 2016; Huang et al., 2017; Howard et al., 2019). The selection of these diverse deep learning models for PE classification was aimed at leveraging their documented success in broad image classification tasks and utilizing their pre-trained weights, a strategy known to enhance performance through the exploitation of learned features from large, diverse datasets (Kim et al., 2022; Zhou et al., 2022; Pan et al., 2023; Ananda et al., 2021). By evaluating their individual performance, the study aimed to assess how their unique architectural traits—ranging from depth and complexity to parameter efficiency and computational demands—perform in PE classification. This comparative analysis facilitated a thorough evaluation of each backbone model's capabilities in addressing challenges like lesion variability and feature extraction accuracy, thereby determining the models most suited for meeting the specificity and sensitivity requirements of PE classification, and providing the foundation for future research and potential clinical applications.

The AG-CNN architecture is fundamentally divided into three core branches: the global branch, the local branch, and the fusion branch. The global branch provides information derived from the global image as input. This branch trains a classification CNN model as its backbone, which includes a series of downsampling blocks, a global pooling layer, and a fully connected (FC) layer with C-dimensions for classification. To normalize the FC layer's output vector, denoted as $pg(c|I)$, a sigmoid layer is added:

$$pg(c|I) = \frac{1}{1 + exp(-p_g(c|I))} \tag{9}$$

here, $I$ symbolizes the global image, and $pg(c|I)$ denotes the probability score of $I$ being classified into the $c^{th}$ class, where $c \in \{1, 2, ..., C\}$ and $C$ represents the total number of classes. The global branch's parameter, $W_g$, is optimized by minimizing the binary cross-entropy (BCE) loss:

$$\mathscr{L}(W_g) = -\frac{1}{C}\sum_{c=1}^{C} l_c log(p_g(c|I)) + (1 - l_c)log(1 - p_g(c|I)) \tag{10}$$

In this equation, $l_c$ is the ground truth label of the $c^{th}$ class.

Meanwhile, the local branch targets the area of embolism, aiming to compensate for the shortcomings associated with exclusively using the global image. Both the global and local branches share the same network structure, but their weights are distinct for their different roles within the framework. The local branch's probability score is represented as $p_j(c|I)$, and $W_l$ denotes the local branch's parameters. The local branch follows the same normalization and optimization processes as the global branch.

The fusion branch is initiated by concatenating the final pooling layers of both the global and local branches. This concatenated layer is then connected to a $C$-dimensional FC layer for final classification, with the probability score denoted as $p_f(c|I,I_c)$. The fusion branch's parameters are denoted as $W_f$, and they are optimized using the same formula as in Eq. (10).

At its core, the AG-CNN architecture is a versatile framework that can be effectively trained with any CNN model as its backbone. This adaptability allowed us to experiment with various backbone architectures, including ResNet, DenseNet, MobileNet, and Inception, to explore their performance within the AG-CNN framework. The simplified visual representation of the AG-CNN architecture in Fig. 4 demonstrates its adaptability, as it can accommodate any classification CNN model as its backbone.

**Detection Framework** For this task, we have chosen state-of-the-art detectors, specifically EfficientDet (Teigen et al., 1995), YOLOv8 (Jocher et al., 2023), and Faster R-CNN (Lee et al., 2011). Representing anchor-based two-stage detectors, Faster R-CNN (Lee et al., 2011) initially proposes potential regions, subsequently refining their locations and classifying them. In contrast, EfficientDet (Teigen et al., 1995), a single-stage detector, directly localizes and categorizes densely proposed anchors. Unlike the previous detectors, YOLOv8 adopts an anchor-free strategy, directly predicting object sizes and centers without using predefined anchor boxes. The purpose of selecting these three diverse models is to ensure that their performances complement each other, leading to more robust and improved results when combined in an ensemble. Three distinct ensembling techniques, namely Weighted Bounding Box Fusion (WBF), Non-Maximum Suppression (NMS), and Non-Maximum Weighted (NMW), were employed to merge the results from the various models. Fig. 5 represents the complete scenario of the detection framework.

The Non-Maximum Suppression (NMS) technique filters out bounding boxes that overlap beyond a certain IoU threshold, usually dropping those with weaker confidence scores (Bodla et al., 2017). From an initial list of detections and their respective scores, the one with the top score is selected, moved to the finalized list, and any overlapping boxes are removed. This procedure continues until all detections are assessed,

$$\text{NMS}(b_1, b_2, t) = \begin{cases} 1 & \text{if } \text{IoU}(b_1, b_2) > t \text{ and } \text{score}(b_1) > \text{score}(b_2) \\ 0 & \text{otherwise} \end{cases} \quad (11)$$

where, $b_1$ and $b_2$ are two bounding boxes, IoU is the Intersection over Union function, $t$ is the IoU threshold, and $score$ is the confidence score function. This equation denotes that we keep the box $b_1$ and suppress the box $b_2$ if they have a high overlap and $b_1$ has a higher score than $b_2$. Otherwise, we do not suppress any box.

Non-maximum weighted (NMW) is a technique for combining predictions of object detection models that uses the IoU value to weigh the boxes. The basic idea is to keep the box with the highest confidence score and compare it with the other boxes (Zhou et al., 2017). If the IoU is greater than a threshold, the other box is weighted by the IoU and added to the final box. Otherwise, the other box is discarded. The procedure for NMW is,

$$\text{NMW}(b_1, b_2, t) = \begin{cases} b_1 + \text{IoU}(b_1, b_2) \cdot b_2 & \text{if } \text{IoU}(b_1, b_2) > t \text{ and } \text{score}(b_1) > \text{score}(b_2) \\ b_1 & \text{otherwise} \end{cases} \quad (12)$$

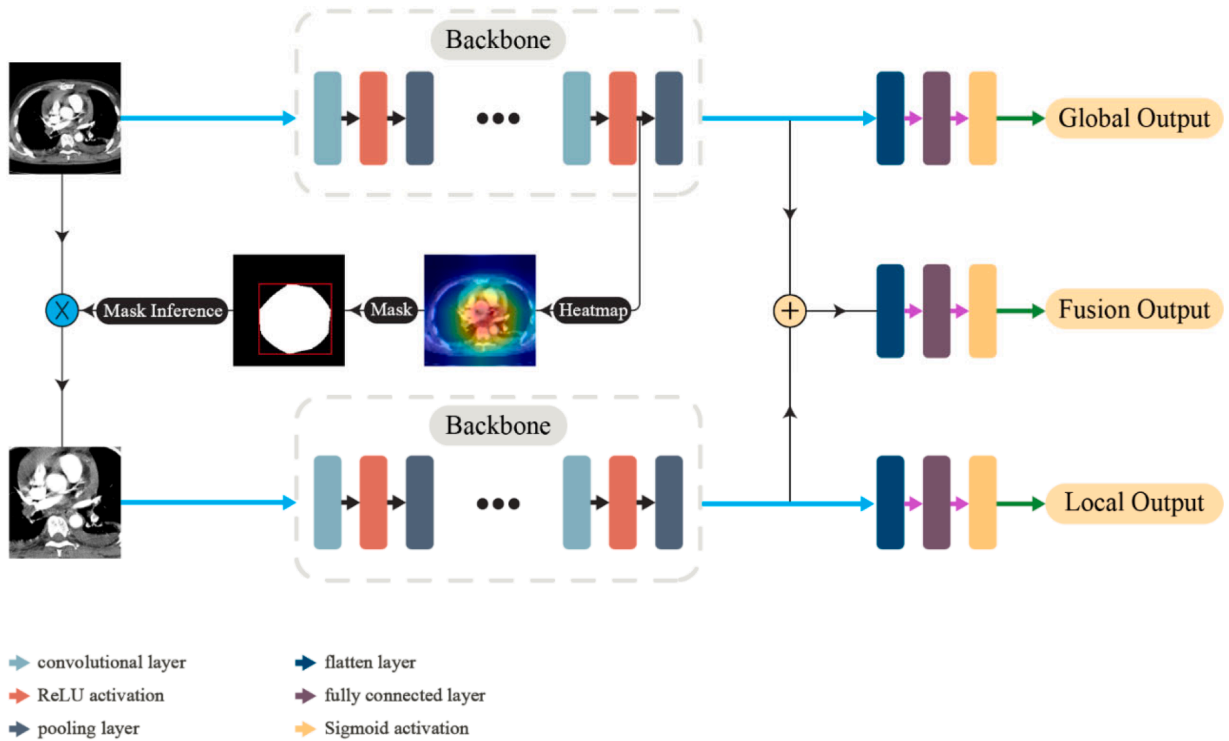

**Fig. 4.** Schematic representation of the AG-CNN architecture.

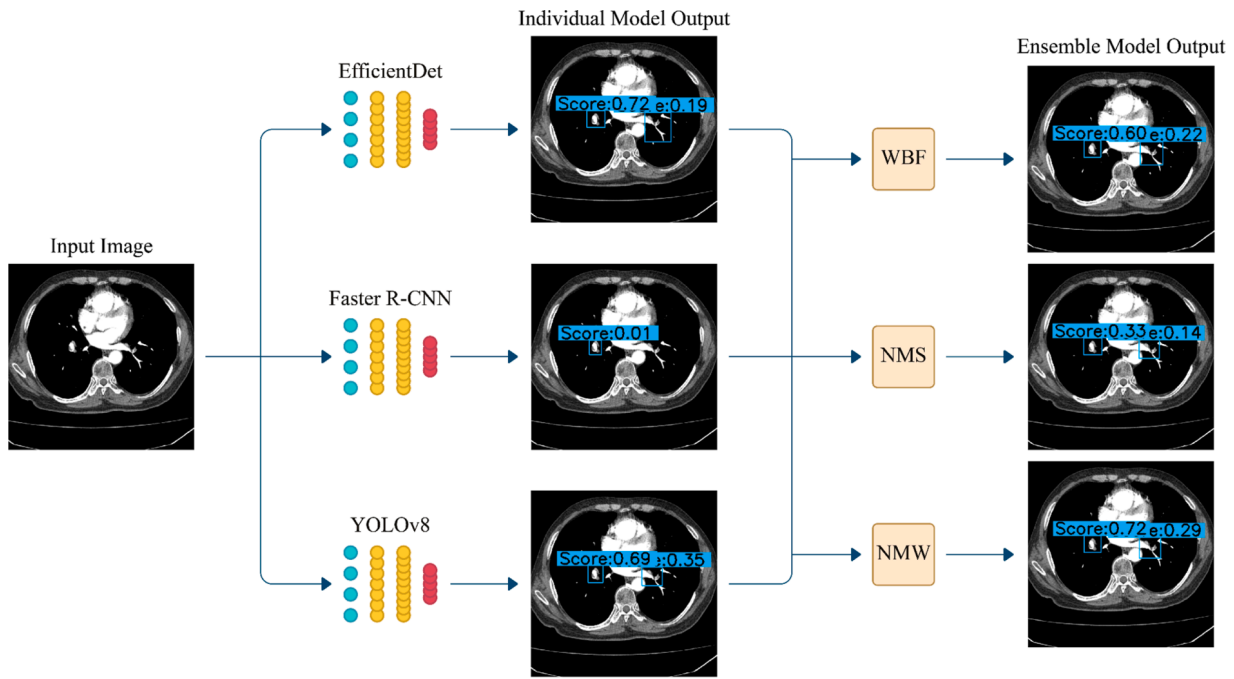

**Fig. 5.** Implementation of various ensembling methods to combine the outputs from different detection models.

In the Weighted Bounding Box Fusion (WBF) method, predictions aren't discarded or diminished. Instead, the method combines the detections by leveraging the confidence scores of all suggested bounding boxes to form averaged boxes (Solovyev et al., 2021). Essentially, boxes are ranked by confidence, and as each box is assessed, it's either merged (by averaging coordinates and scores) with a previously fused box if their overlap exceeds a threshold or added to the fused box list if it doesn't. The procedure for WBF is as follows,

$$\text{WBF}(b_1, b_2, t) = \begin{cases} \dfrac{\text{score}(b_1) \cdot b_1 + \text{score}(b_2) \cdot b_2}{\text{score}(b_1) + \text{score}(b_2)} & \text{if } \text{IoU}(b_1, b_2) > t \\ [b_1, b_2] & \text{otherwise} \end{cases} \quad (13)$$

**Classifier-Guided Detection** For an input image, referred to as $I$, the fusion branch classifier's output, denoted as $pf(embolism|I)$, indicates the likelihood of the image belonging to the PE class. To enhance our embolism detector's accuracy, we propose a heuristic combination strategy with the classifier's output. If an image $I$ has a prediction $pf(embolism|I) \geq \theta$, where $\theta$ represents the threshold for binary classification, we keep all detected embolisms. However, if $pf(embolism|I) < \theta$, only bounding boxes with a confidence score above 0.018 are considered. The threshold was chosen through a heuristic approach on the validation dataset to ensure that it does not compromise the detection sensitivity while optimizing precision. The proposed framework is illustrated in Fig. 6.

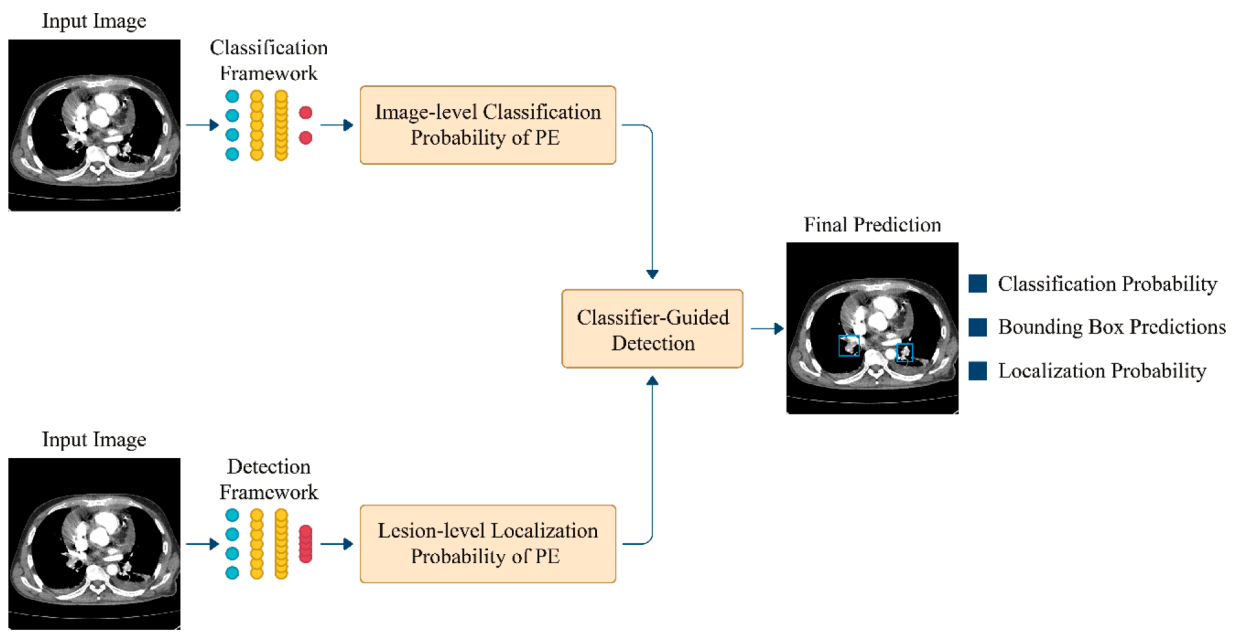

**Fig. 6.** Schematic diagram of the Classifier-Guided Detection approach. The classification framework takes a CTA image slice as input and predicts its image-level probability of containing PE. The detection framework provides bounding boxes along with probability scores of PE at the lesion-level for the same image slice. The Classifier-Guided Detection mechanism then combines the two outputs and maximizes the detection performance.

*3.5. Evaluation metrics*

This study evaluated a range of metrics to determine the model's predictive capacity. The performance of the classification task is evaluated and presented using various statistical measures, including the Area Under the Receiver Operating Characteristic (AUROC), accuracy, precision, sensitivity, specificity, and the F1 score. These metrics can be described as follows,

$$Accuracy = \frac{TP + TN}{TP + TN + FP + FN} \quad (1)$$

$$Precision = \frac{TP}{TP + FP} \quad (2)$$

$$Sensitivity = \frac{TP}{TP + FN} \quad (3)$$

$$Specificity = \frac{TN}{TN + FP} \quad (4)$$

where, TP, TN, FP, and FN represent true positive, true negative, false positive, and false negative, respectively.

The F1 score is a measure that combines precision and recall into a single metric to provide a balanced evaluation,

$$F1Score = 2*\frac{Precision*Recall}{Precision + Recall} \quad (5)$$

AUROC is used for measuring the ability of a model to distinguish between positive and negative instances across different probability thresholds (Draelos, (2019, 14–8-2023).). It is calculated as the area under the ROC curve,

$$AUROC = \int_0^1 TPR(FPR^{-1}(t))dt \quad (6)$$

where *TPR* is the true positive rate and *FPR* is the false positive rate.

In evaluating the performance of the detection task, the study employs the COCO standard, using the mean average precision (*mAP*) as a key metric, alongside the statistical measures previously outlined (Padilla et al., 2020). A prediction is considered a true positive if it overlaps with a ground truth lesion of the same class, achieving an IoU greater than the threshold set. *mAP* quantifies the average of the precision values at different recall levels,

$$mAP = \frac{1}{n}\sum_{i=1}^{n} AP_i \quad (7)$$

where $n$ is the number of classes and $AP_i$ is the average precision for class *i*. AP is computed as the area under the precision-recall curve,

$$AP = \int_0^1 p(r)dr \quad (8)$$

where $p(r)$ is the precision at recall *r*.

## 4. Results and discussion

In this section, we initially assess the individual performance of classification and detection frameworks in identifying pulmonary embolisms. Subsequently, we demonstrate the effectiveness of the classifier-guided detection approach achieved through the strategic integration of these two frameworks.

*4.1. Classification model evaluation*

We evaluate the performance of different models as backbones on the test dataset which is summarized in Table 2. We specifically present the results for the PE class without including the weighted average of the Non-PE class, as it better reflects the model's effectiveness in classifying PE. For training, we utilized pre-trained weights from the ImageNet dataset (Deng et al., 2009) and implemented the proposed framework using PyTorch. All the branches of the network were optimized using Adam with a Cosine Annealing Learning Rate Scheduler using an initial learning rate of 1e-7, 1e-7 and 1e-5 for the global, local and fusion branches respectively. To train the network, we used the Binary Cross Entropy (BCE) loss function.

**Baseline Performance and Local Branch Performance Assessment** We present the results of the baseline, specifically the global branch, with ResNet-50, ResNet-152, DenseNet-121, DenseNet-201, MobileNetV3-Large, and Inception-v3 models as backbone architectures in Table 2. Subsequently, we assess the performance of the corresponding local branches, which aim to enhance the classification accuracy by training on the cropped discriminative patches to provide supplementary attention mechanisms in conjunction with the global branch. Evidently, the AUROC scores of the local branches exhibit a decrease compared to the values achieved by their corresponding global branch models. This divergence may be attributed to the process of estimating and cropping the lesion region which may result in information loss, affecting the local branch's performance accuracy. It is noteworthy that the performance drop in the local branch is less severe for ResNet-152, which has the highest number of model parameters among the six models. Conversely, the performance drop is most marked for MobileNetV3-Large, showing a 7.7 % reduction in AUROC score compared to its global branch counterpart. This trend highlights the relationship between a neural network's capacity, influenced by its parameter count, and its performance. ResNet-152, as a deeper model, effectively learns complex, hierarchical features, making it more invariant to local changes in the input space. In contrast, the lightweight architecture of MobileNetV3-Large, while efficient, may lack the depth required to adapt to such variations, resulting in a more significant performance decrease.

**Table 2**
Comparison of Model Performance on Test Set.

| Backbone Model | Parameters | Method | Accuracy | Precision | Sensitivity | Specificity | F1-Score | AUROC |
|---|---|---|---|---|---|---|---|---|
| ResNet-50 | 25.56 M | Global Branch | 0.818 | 0.640 | 0.904 | 0.781 | 0.749 | 0.894 |
| | | Local Branch | 0.803 | 0.618 | 0.916 | 0.755 | 0.738 | 0.881 |
| | | Fusion Branch | 0.842 | 0.682 | 0.902 | 0.818 | 0.777 | 0.909 |
| ResNet-152 | 60.19 M | Global Branch | 0.817 | 0.633 | 0.932 | 0.767 | 0.754 | 0.900 |
| | | Local Branch | 0.834 | 0.668 | 0.894 | 0.807 | 0.764 | 0.890 |
| | | Fusion Branch | 0.833 | 0.661 | 0.916 | 0.797 | 0.768 | 0.901 |
| DenseNet-121 | 7.98 M | Global Branch | 0.800 | 0.733 | 0.532 | 0.864 | 0.617 | 0.845 |
| | | Local Branch | 0.775 | 0.614 | 0.684 | 0.813 | 0.647 | 0.814 |
| | | Fusion Branch | 0.860 | 0.728 | 0.852 | 0.863 | 0.785 | 0.926 |
| DenseNet-201 | 20.01 M | Global Branch | 0.833 | 0.704 | 0.768 | 0.861 | 0.735 | 0.896 |
| | | Local Branch | 0.795 | 0.637 | 0.739 | 0.817 | 0.684 | 0.837 |
| | | Fusion Branch | 0.850 | 0.707 | 0.871 | 0.844 | 0.780 | 0.920 |
| MobileNetV3-Large | 5.48 M | Global Branch | 0.803 | 0.634 | 0.824 | 0.794 | 0.716 | 0.883 |
| | | Local Branch | 0.683 | 0.487 | 0.859 | 0.609 | 0.621 | 0.806 |
| | | Fusion Branch | 0.802 | 0.635 | 0.852 | 0.788 | 0.727 | 0.888 |
| Inception-v3 | 27.16 M | Global Branch | 0.865 | 0.748 | 0.831 | 0.879 | 0.787 | 0.923 |
| | | Local Branch | 0.829 | 0.707 | 0.749 | 0.866 | 0.727 | 0.878 |
| | | Fusion Branch | 0.874 | 0.754 | 0.862 | 0.879 | **0.805** | **0.927** |

Furthermore, both the global and local branches with DenseNet-121 as the backbone architecture demonstrate low sensitivity compared to other models. This might be attributed to the fundamental design of DenseNets, which prioritizes parameter efficiency (Huang et al., 2017). While this efficiency is advantageous for minimizing overfitting and reducing computational demands, it may limit the model's capacity to accurately capture and model the complex patterns necessary for identifying all true positives, leading to lower sensitivity in DenseNet-121. For qualitative analysis of the global context, we provide representations of the Class Activation Map (CAM) for the Inception-v3 model on the dataset in Fig. 7. Notably, these CAMs not only add "explainability" to the model but also suggest a consistent correspondence between the attention-focused regions and the corresponding lesion areas present in the CTA images through visualization.

**Performance Analysis of Integrating Global and Local Branches**

The effectiveness of the fusion branch, responsible for producing the ultimate classification metrics, is demonstrated in Table 2. Notably, the fusion branch outperforms both the corresponding global and local branches for all state-of-the-art models used as backbones in this study. For instance, when utilizing DenseNet-121 as the backbone, the AUROC metric of the fusion branch outperforms the AUROC values of global and local branches by 8.1 % and 11.2 %, respectively, which is a significant performance leap. In Fig. 8, we present the ROC curves for the PE class with various backbone models which demonstrate consistently improved performance of the fusion branch compared to the global and local branches across all these models. As depicted in Table 2, AG-CNN with Inception-v3 as backbone yields both the highest AUROC score of 0.927 and the highest F1-score of 0.805 compared to other models. The enhanced performance of Inception-v3 in the fusion branch likely stems from its adeptness in handling scale variance (Wang and Breckon, 2022). This capability is particularly beneficial in the context of pulmonary embolisms, where the dataset encompasses a wide range of embolism sizes, including Saddle, Lobar, Segmental, and Sub-segmental Embolisms (Choi et al., 2014). Inception-v3's multi-scale feature capture effectively adjusts to these diverse embolism sizes, contributing significantly to its improved AUROC and F1 scores compared to the other models.

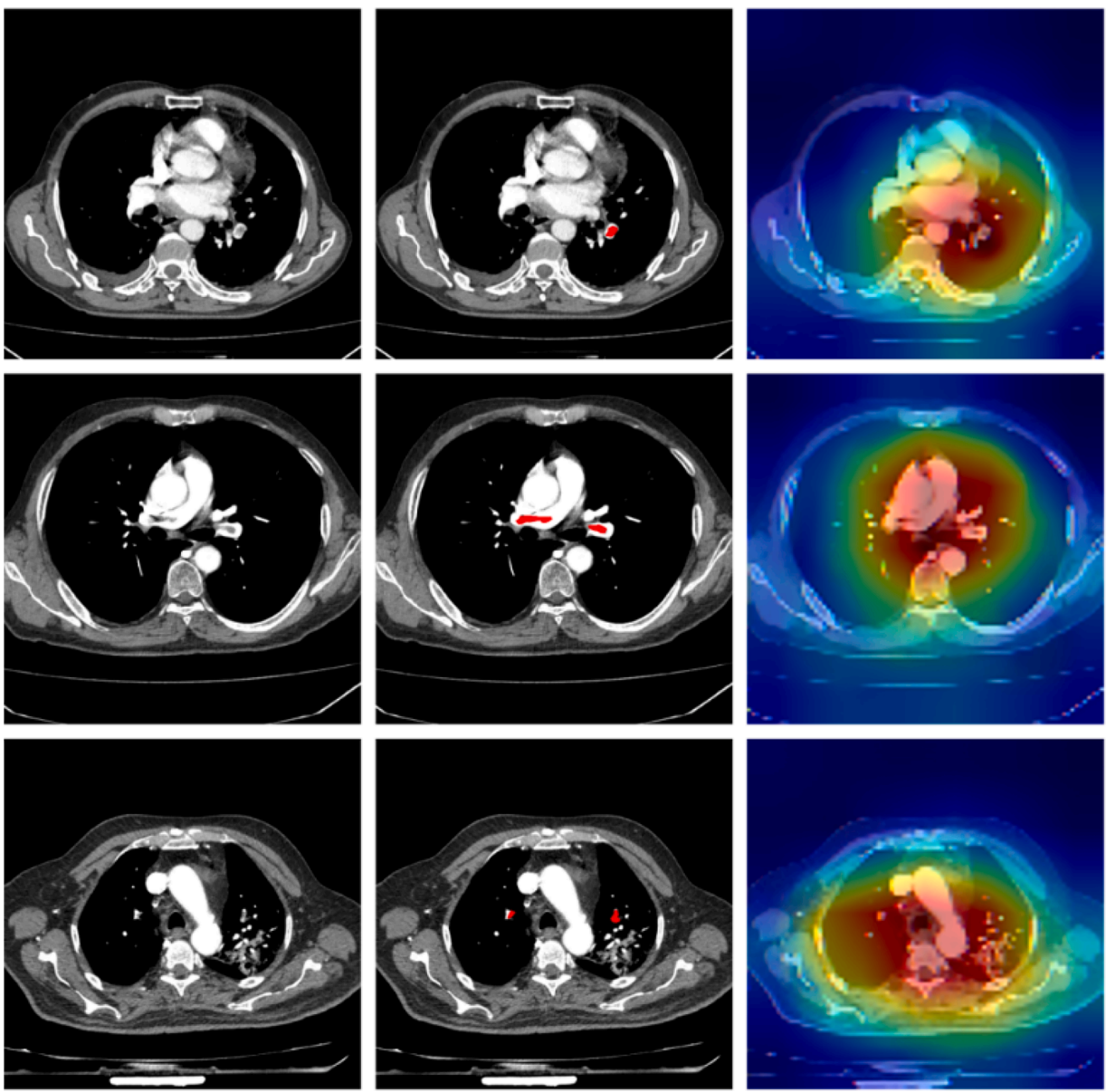

**Fig. 7.** Class activation map (CAM) representation of the CTA image slices (Left) original CTA images. (Middle) corresponding images containing red-marked regions for visual representations of the lesion areas. Note that we do not use images marked with red regions for training or testing. (Right) corresponding CAM images for the Inception-v3 model as the backbone. Higher/lower attention is denoted with red/blue.

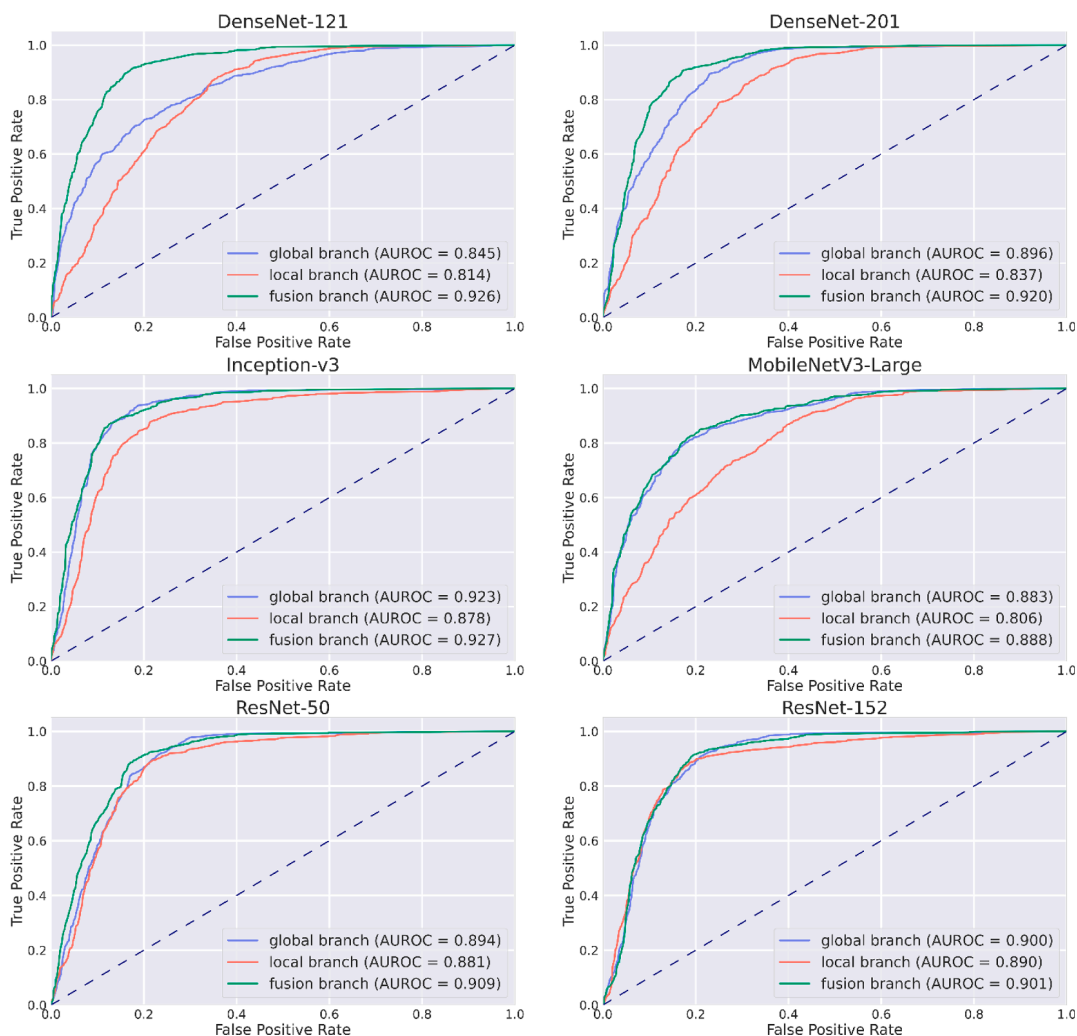

**Fig. 8.** ROC curves of AG-CNN on PE classification for various models as backbones.

As anticipated, the precision metric is impacted by the disparity in the distribution of PE and Non-PE classes, with a relatively small number of slices containing PE (approximately 26 %) in the dataset leading to lower precision values. Nevertheless, in the clinical context, prioritizing the sensitivity of positive cases holds greater significance than precision, even if it results in more false positives. This approach aligns with AG-CNN's intended purpose as a diagnostic aid, where the model predictions can be subsequently reviewed and filtered by radiologists. Such collaboration between human experts and AI is expected to enhance both precision and sensitivity, ultimately contributing to improved diagnosis.

### 4.2. Detection model evaluation

Pretrained weights from the MS COCO dataset (Lin et al., 2014) were used to train each model on the training set. The YOLOv8 models, developed in PyTorch, were optimized utilizing the Adam optimizer in conjunction with a Step Learning Rate Scheduler. Meanwhile, the Faster R-CNN Resnet-101 and EfficientDet-D0 models were deployed using Tensorflow 2.0 and optimized through Stochastic Gradient Descent (SGD) with a Cosine Decay Learning Rate Scheduler using a base learning rate of 1e-3.

In this study, the ground truth bounding boxes were deliberately extended to provide the model with arterial context. This, however, necessitated evaluating the model's *mAP* metric at a reduced IoU threshold of 0.2 compared to the conventional *mAP* at IoU of 0.5, as the threshold of 0.2 represents the model's performance more genuinely. The comparison between ground truth annotations and predicted bounding boxes on a sample image is demonstrated in Fig. 9. Despite the lower overlap with the ground truth, the visual analysis showed that the models accurately enclosed the embolisms within their predicted bounding boxes. However, the model's performance at $mAP_{50}$ has also been provided alongside $mAP_{20}$. The following sections will reveal a marginal difference between results at $mAP_{20}$ and $mAP_{50}$. This negligible gap further substantiates our claim, suggesting that the results at the set threshold of 0.2 aren't inflated by irrelevant predictions, thus reinforcing our stance that $mAP_{20}$ results aptly represent the model's true capabilities.

**Individual Model Performance** Table 3 shows the quantitative results of different versions of YOLOv8 models along with the Faster R-CNN Resnet-101 and EfficientDet-D0 models. EfficientDet-D0 surpasses other models in detecting PE with the highest $mAP_{20}$ score of 0.820 and the highest sensitivity of 0.886. Meanwhile, YOLOv8l ranks second, achieving a $mAP_{20}$ score of 0.794 and sensitivity of 0.872 outperforming its other versions, YOLOv8s and YOLOv8m.

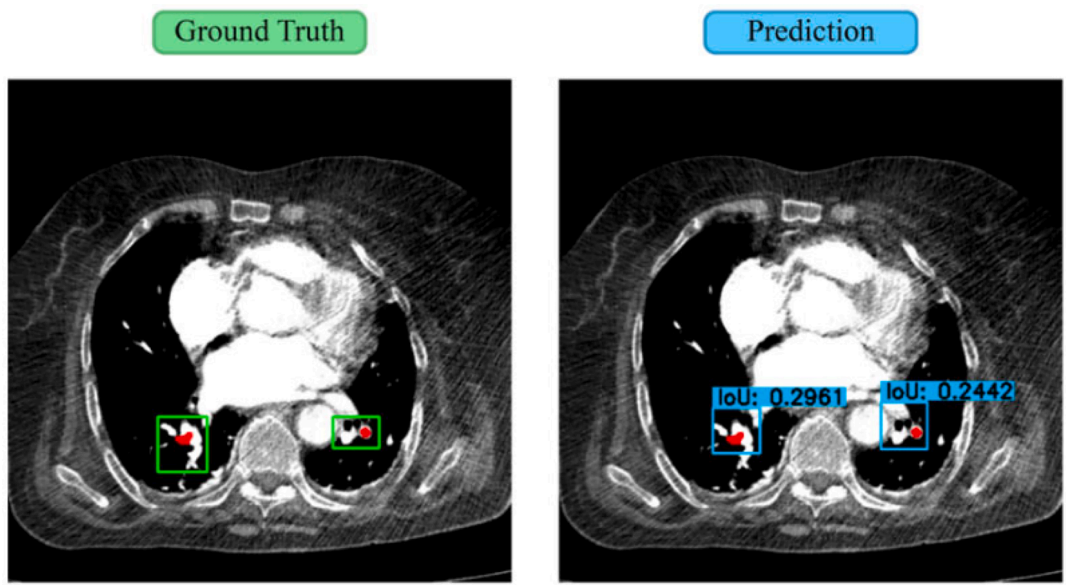

**Fig. 9.** Comparison of ground truth annotations and predicted bounding boxes with IoU overlap for PE detection.

**Table 3**
Detection Model Performance on Test Set.

| Model | Precision | Sensitivity | F1-Score | $mAP_{20}$ | $mAP_{50}$ |
| --- | --- | --- | --- | --- | --- |
| EfficientDet-D0 | 0.447 | **0.886** | 0.594 | **0.820** | **0.791** |
| Faster R-CNN ResNet-101 | 0.689 | 0.779 | 0.731 | 0.668 | 0.638 |
| YOLOv8l | 0.672 | 0.872 | 0.759 | 0.794 | 0.769 |
| YOLOv8m | 0.739 | 0.835 | 0.784 | 0.752 | 0.729 |
| YOLOv8s | 0.673 | 0.836 | 0.746 | 0.739 | 0.707 |

Individual models' predictions along with the ground truth annotations are shown in Fig. 10 for three sample images so qualitative results can be inspected. EfficientDet-D0, given its higher sensitivity, predicts more false positives evident in the middle image, explaining its reduced precision score. For the rightmost image, while most models can correctly detect the two ground truth lesion areas, Faster R-CNN ResNet-101 fails to detect one of the embolisms, representing its lower sensitivity in comparison to the other models.

**Ensemble Model Performance** For the model ensembling operation, we combined the predictions of EfficientDet-D0, Faster R-CNN Resnet-101, and the top-performing YOLO variant which is YOLOv8l. The results were generated using three different ensembling methods, specifically Non-Maximum Suppression (NMS), Weighted Box Fusion (WBF), and Non-Maximum Weighted (NMW). For employing the ensembling techniques, we chose an IoU threshold of 0.3 in this experiment. A weighted score of 3.0, 2.5, and 1.0 was used for the EfficientDet-D0, YOLOv8l, and Faster R-CNN Resnet-101 models respectively to perform the ensembling operation. The weight distribution was based on the individual model's performance, focusing on their $mAP$ scores with the best-performing model being assigned the most weight and the worst performing model being assigned the least weight. Additionally, the probability scores from each model's predictions were factored into this weighting strategy for the ensemble operation. The EfficientDet-D0 model, due to its higher sensitivity, can detect some embolisms that the other two models cannot, although with a low probability score. The decision fusion operation would have discarded such predictions due to their lower probability score. By assigning a weighted score of 3.0 to the EfficientDet-D0 model, these low-probability predictions were equivalently scaled against predictions from other models, enabling them to surpass the probability threshold set during the decision fusion process. The performance metrics after ensembling are presented in Table 4.

In all instances, the ensemble models show a significant improvement over individual detection models in both $mAP$ scores and sensitivity. The WBF-based ensemble model reaches the highest $mAP$ of 0.867, marking a 4.7 % increase compared to the top-performing detection model, EfficientDet-D0. Furthermore, the NMW-based ensemble model achieves the highest sensitivity of 0.927, improving by 4.1 % over the most sensitive individual model, also EfficientDet-D0. The NMS method suffers from lower sensitivity scores compared to the other ensemble methods. This might be attributed to the NMS ensemble operation which eliminates overlapping bounding boxes leading to miss embolisms that are close together (Bodla et al., 2017). On the other hand, the NMW method, by adjusting the highest-scoring box rather than discarding overlapping predictions, tends to preserve more true positives (Zhou et al., 2017). However, this approach may also inadvertently retain false positives, explaining the method's increased sensitivity but at the cost of reduced precision, as reflected in the performance metrics. Overall, the WBF-based ensemble method demonstrates the most balanced performance, effectively optimizing precision and sensitivity, as evidenced by its leading F1-score of 0.777.

Fig. 11 shows a visual comparison of performance between the ensembled outputs and the three fundamental models on an individual test image. When contrasting single-model outcomes with ensemble-derived results, it's evident that ensembling offers superior performance. For instance, ensembling can mitigate the case where a detection model misses a lesion area, as illustrated in the figure.

**Effect of the Classifier-Guided Detection Operation** For the classifier-guided detection step, we chose the AG-CNN fusion branch with Inception-v3 backbone as the optimal classifier for this task. This classifier has been selected due to its higher specificity. A higher specificity ensures that the model accurately identifies negative cases, reducing FPs. For the image slices predicted as positive by the classifier for containing embolism, all detections are retained and for other image slices, only detections with a probability score exceeding the threshold of 0.018 are kept. As previously mentioned, we set the specific threshold value of 0.018 based on heuristic evaluations on the validation dataset, with the primary aim of enhancing the precision while preserving the detection sensitivity. Selecting a threshold value greater than 0.018 resulted in the exclusion of some TP predictions, thereby reducing sensitivity. Conversely, thresholds below 0.018 retained some FP predictions leading to an unoptimized precision value. Given these considerations, a threshold of 0.018 was the optimal choice for our method. Fig. 12 represents the effectiveness of the Classifier-guided framework in reducing the false positives from the ensemble-based detections. As shown in the figure, the proposition effectively removes the false positives retaining only the relevant embolisms. A comparison of the detection performance between the ensemble models and the classifier-guided framework is provided in Table 5. We observe that the proposed framework, which influences the detection performance through the classifier's predictions, leads to improved performance. Since only false positives are getting reduced, the sensitivity remains the same but the precision increases. Consequently, the proposed framework yields improved *mAP* and F1 scores than the ensemble models.

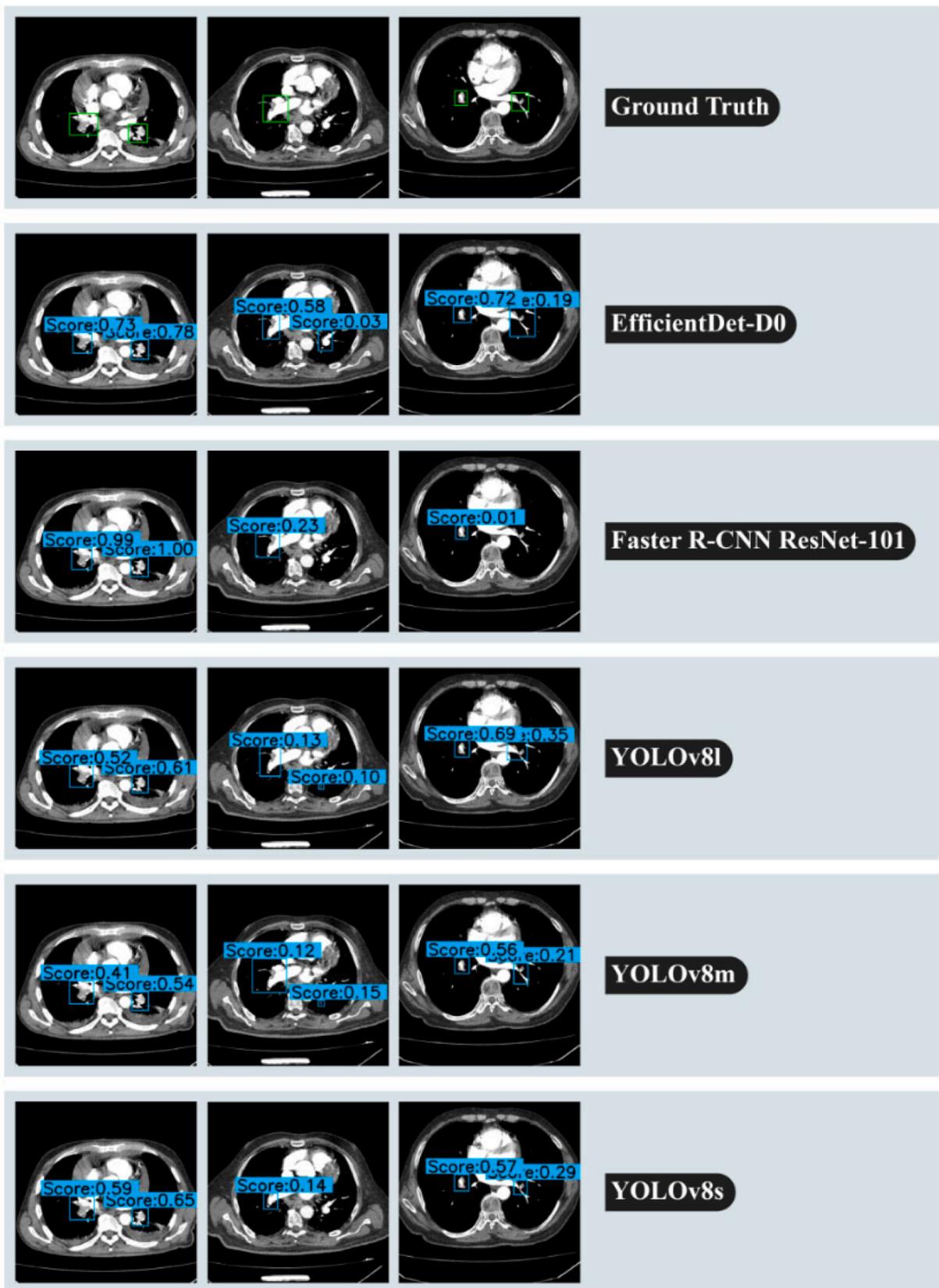

**Fig. 10.** Performance of individual detection models on sample test images.

**Table 4**
Ensemble Method Performance for various methods on Test Set.

| Method | Precision | Sensitivity | F1-Score | $mAP_{20}$ | $mAP_{50}$ |
|---|---|---|---|---|---|
| WBF-based ensemble | 0.683 | 0.901 | 0.777 | **0.867** | **0.845** |
| NMS-based ensemble | 0.699 | 0.870 | 0.775 | 0.863 | 0.840 |
| NMW-based ensemble | 0.566 | **0.927** | 0.703 | 0.864 | 0.840 |

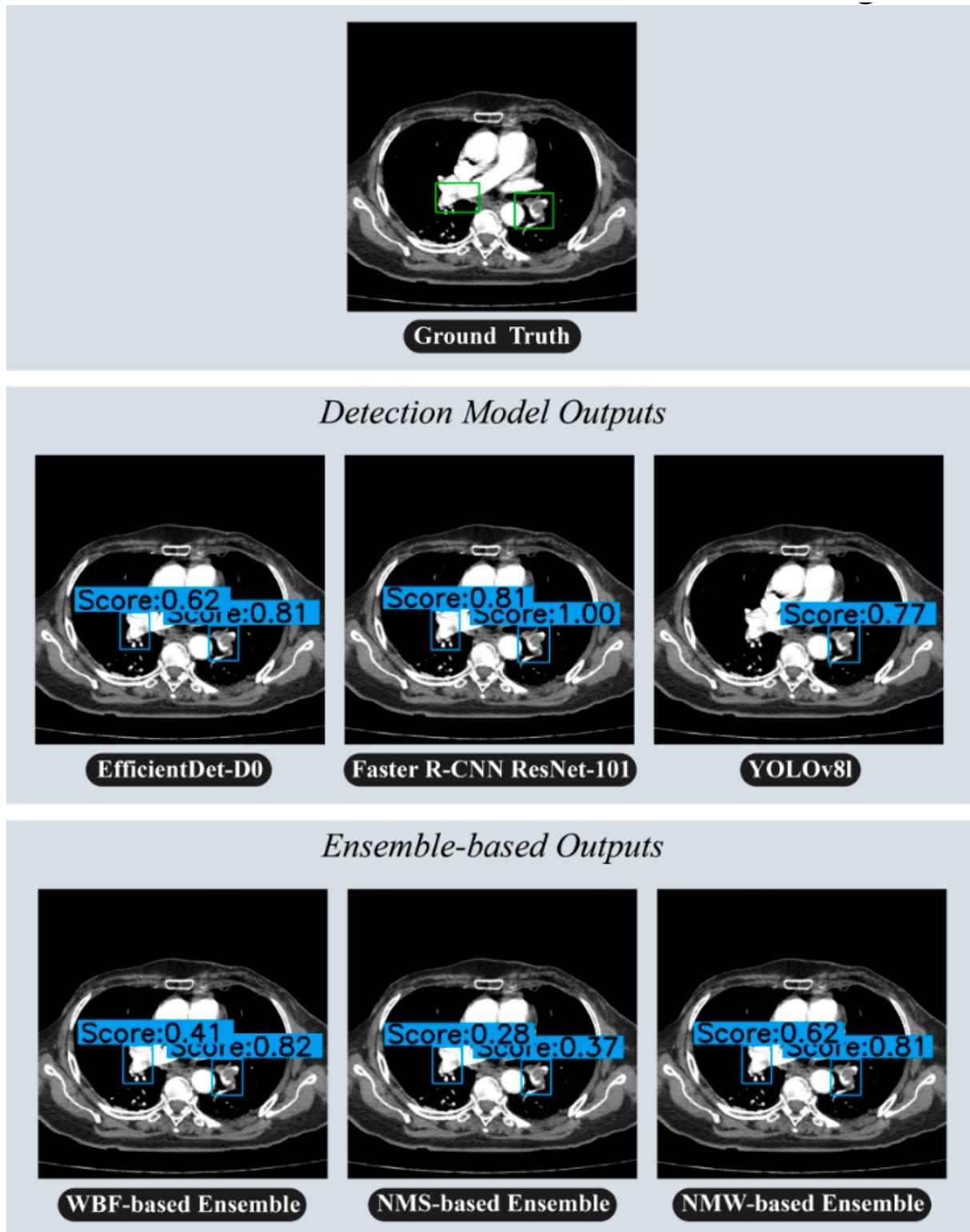

**Fig. 11.** Performance comparison between individual detection models and the ensembled outputs on a sample test image.

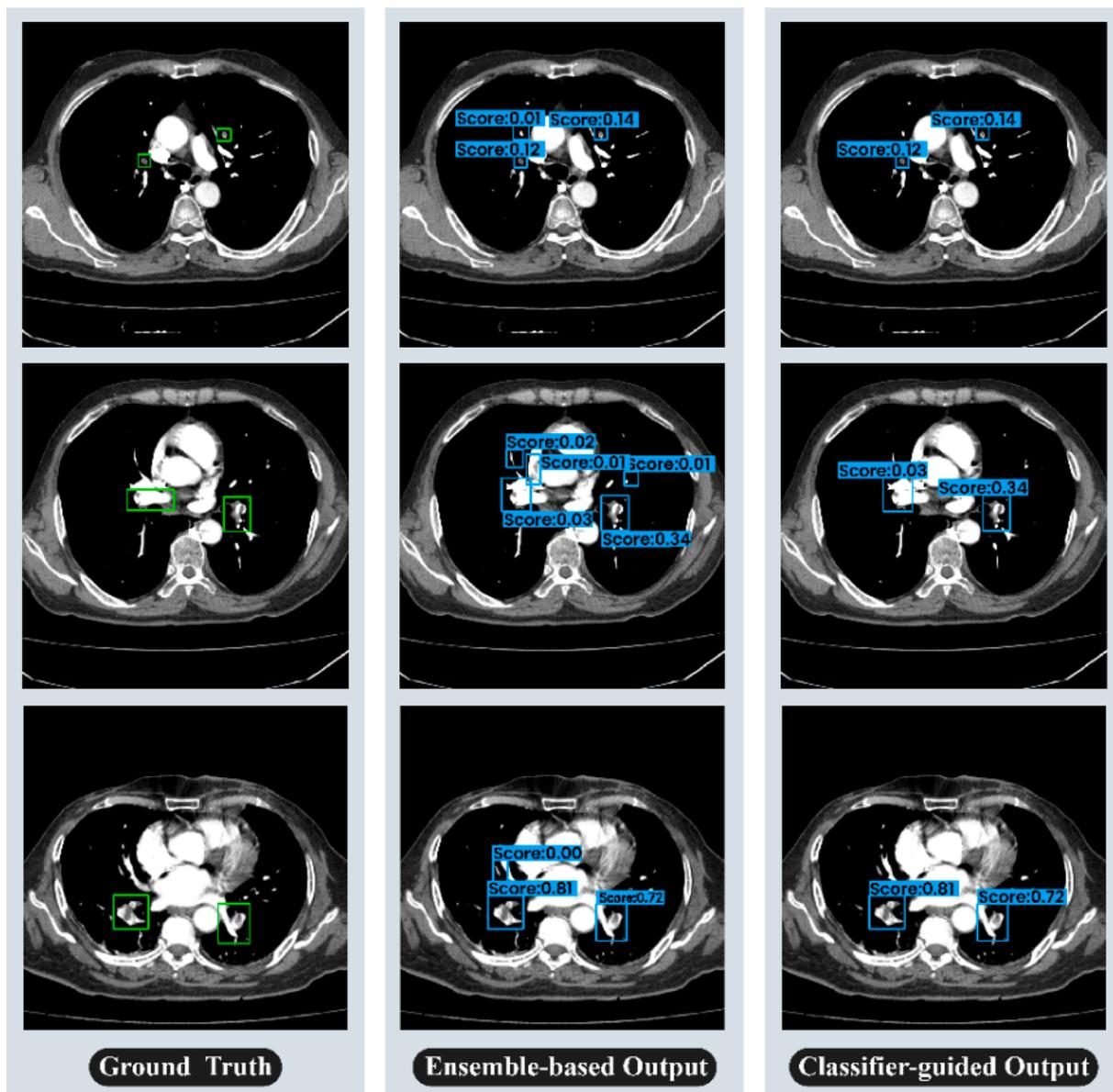

**Fig. 12.** Sample images representing the effect of the classifier-guided framework on the ensemble-based outputs leading to the False Positive (FP) reduction. The Classifier-guided framework leads to the reduction of one FP from the WBF-based output (top), three FPs from the NMS-based output (middle), and one FP from the NMW-based output (bottom).

**Table 5**
Ensemble Method Performance for various methods on Test Set.

| Method | Precision | Sensitivity | F1-Score | $mAP_{20}$ | $mAP_{50}$ |
| --- | --- | --- | --- | --- | --- |
| WBF-based output | 0.683 | 0.901 | 0.777 | 0.867 | 0.845 |
| Classifier-guided WBF-based output | 0.687 | 0.901 | 0.779 | **0.869** | **0.846** |
| NMS-based output | 0.699 | 0.870 | 0.775 | 0.863 | 0.840 |
| Classifier-guided NMS-based output | 0.703 | 0.870 | 0.777 | 0.865 | 0.841 |
| NMW-based output | 0.566 | 0.927 | 0.703 | 0.864 | 0.840 |
| Classifier-guided NMW-based output | 0.570 | 0.927 | 0.705 | 0.865 | 0.841 |

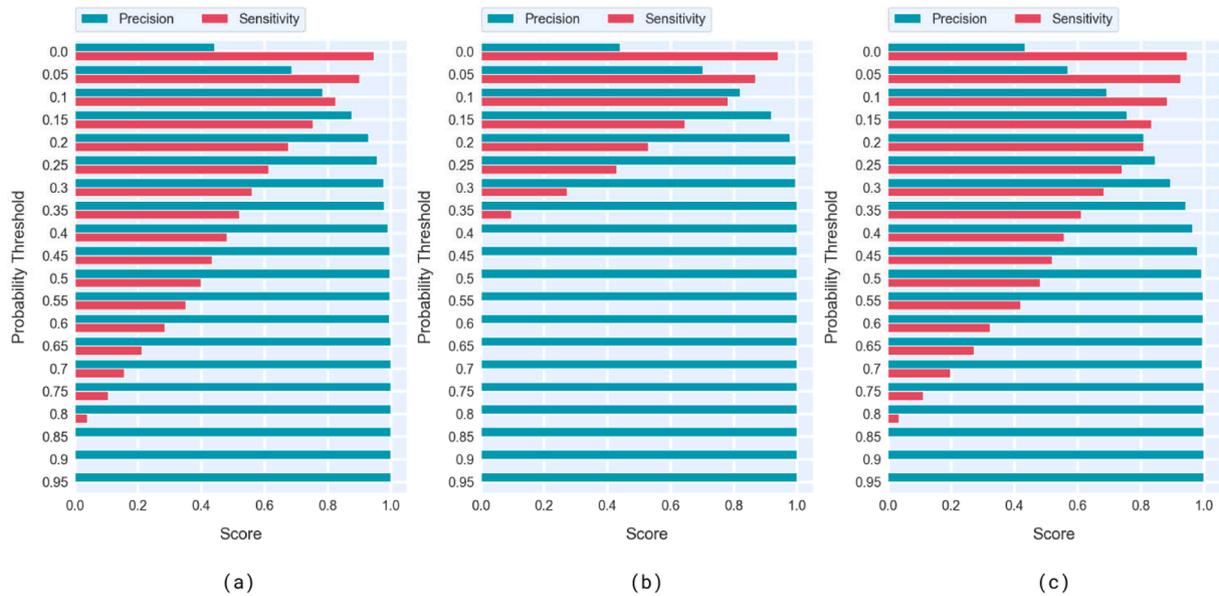

**Fig. 13.** Comparison of precision and sensitivity across different probability thresholds of the classifier-guided framework over the (a) WBF-based ensemble outputs, (b) NMS-based ensemble outputs, and (c) NMW-based ensemble outputs.

The proposed framework, which integrates predictions from both the classifier and ensemble-based models, attains *mAP* scores of 0.869 at an IoU threshold of 0.2 and 0.846 at an IoU threshold of 0.5. Notably, when considering the IoU threshold at 0.5, our model outperforms the prevailing leaderboard on the FUMPE dataset by 3.7 % (Long et al., 2021). However, the IoU threshold of 0.2 more accurately reflects the true capabilities of our model.

The precision across all three outputs is notably lower, primarily because we have deliberately set a low probability threshold at $p \geq 0.005$. This setting aims to achieve the highest possible sensitivity, a critical measure in medical diagnosis. As illustrated in Fig. 13, different operating points showcase the trade-off between precision and sensitivity in our detection framework. Using the conventional probability threshold of 0.5 yields reduced sensitivity for our detection framework. However, in clinical applications, the emphasis often shifts towards maximizing sensitivity to critically minimize the rate of false negatives. While our classifier-guided approach built atop the NMW-based output attains the maximum sensitivity, it also introduces a higher rate of false positives. This rise in false positives subsequently diminishes both precision and the *mAP* score, especially when compared to its WBF-based counterpart. A similar trend is observable in Fig. 13, where the precision scores for the classifier-guided structure over the NMW-based output (Fig. 13 (c)) consistently lag behind the other two frameworks. Nonetheless, our approach offers a thoughtful balance, particularly in clinical contexts where the emphasis on sensitivity is crucial while maintaining a keen awareness of the trade-offs involved.

### 4.3. Comparison with state-of-the-art

We compare our classification and detection results with the state-of-the-art methods (Long et al., 2021; Xu et al., 2023; Khachnaoui et al., 2022; Islam et al., 2024) on the FUMPE dataset in Table 6.

Compared to the existing methods, our proposed AG-CNN model establishes a new benchmark in the domain of PE classification, achieving a high sensitivity of 0.862. The enhanced sensitivity indicates a superior ability of our proposed algorithm to correctly identify true positives, a crucial attribute in medical imaging diagnostics where missing a true case can have serious repercussions. Although the study presented by Khachnaoui et al. reports a precision value of 1.6 % greater than our proposed method, their approach suffers from a much lower sensitivity than ours. Our model's notable balance between precision and sensitivity results in a leading F1 score of 0.805, which is 13.9 % higher than their reported score, illustrating an optimized trade-off crucial for clinical diagnostics. Notably, our classification model attains better performance despite the dataset constraints; being trained on the smaller FUMPE dataset in contrast to Islam et al.'s model, which utilized the larger RSNA PE dataset and used the FUMPE dataset for external validation. The observed generalizability and robust performance of AG-CNN, despite the dataset size, suggest a significant advancement in the model's ability to generalize from limited data, an essential characteristic for scalable machine learning solutions.

**Table 6**
Comparison with state-of-the-art methods.

| Approach | Reference | Dataset | Method | Performance |
|---|---|---|---|---|
| Classification | Khachnaoui et al. (Khachnaoui et al., 2022) | FUMPE (644 images) | VGG-16, Inception-v3 | Precision: 0.770<br>Sensitivity: 0.587<br>Specificity: 0.667<br>F1 Score: 0.666 |
| | Islam et al. (Islam et al., 2024) | RSNA PE (1,790,624 images)<br>FUMPE (external validation on 8792 images) | SeXception | Sensitivity: 0.839<br>Specificity: 0.839 |
| | **Proposed Method** | **FUMPE (8792 images)** | **AG-CNN** | **Precision: 0.754**<br>**Sensitivity: 0.862**<br>**Specificity: 0.879**<br>**F1 Score: 0.805** |
| Detection | Long et al. (Long et al., 2021) | FUMPE (8792 images) | Probability-based Mask R-CNN | $mAP_{50}$: 0.809 |
| | Xu et al. (Xu et al., 2023) | Tianjin (71,488 images)<br>FUMPE (external validation on 8792 images) | YOLOv4 | $mAP_{50}$: 0.727<br>Precision: 0.448<br>Sensitivity: 0.841 |
| | **Proposed Method** | **FUMPE (8792 images)** | **Classifier-guided Detection** | $mAP_{50}$: **0.846**<br>$mAP_{20}$: **0.869**<br>**Precision: 0.687**<br>**Sensitivity: 0.901**<br>**F1-Score: 0.779** |

Our proposed classifier-guided detection approach which integrates predictions from both the classifier and ensemble-based models contributes new state-of-the-art to the community: $mAP_{50}$ of 0.846. It outperforms the former benchmark set by Long et al.'s probability-based Mask R-CNN on the FUMPE dataset by a notable improvement of 3.7 %. Moreover, our detection method not only surpasses Xu et al.'s *mAP* score but also excels in maintaining both high sensitivity and precision, a feat often challenging as higher sensitivity can typically compromise precision. This dual enhancement is indicative of the sophisticated balance our method achieves in providing a robust and dependable tool for medical image analysis.

*4.4. Limitation and future research*

While the present work has achieved a promising performance in detecting PE, it still has important limitations,

- The study used a dataset that provided patient data regarding only the PE condition in the lungs. However, in real life, patients may have conditions other than just PE in their lungs in which case our model may confuse PE with other pathologies. Therefore, our framework's ability to accurately detect PE amidst other diseases could not be evaluated.
- In our study, the models deployed don't utilize information from preceding or subsequent image slices when identifying PE. This occasionally leads to missed embolisms in certain slices, an issue that might have been mitigated if the models had considered context from adjacent slices.

The above limitations could be addressed in our future works,

- We will expand our study to include broader datasets that encompass a wider range of lung conditions. This approach will refine the model's ability to differentiate PE from other pathologies and enhance its generalizability and applicability across different scenarios.
- A potential design for the model could involve taking spatial context from multiple consecutive slices as separate inputs, processing each slice with shared CNN layers and then merging the features before making a prediction. This way, the model can utilize context from adjacent slices. Alternatively, leveraging 3D-CNNs can be effective, as they process a small volume (several consecutive slices) as input and hence can learn spatial patterns across slices.
- The future research scope of our study will also incorporate the segmentation of PE along with classification and detection utilizing diverse datasets, ensuring its suitability for real-time deployment in the field of medical diagnosis.

## 5. Conclusion

The timely and accurate diagnosis of PE remains of paramount importance in medical imaging, given the serious repercussions associated with the condition. This study delved deeply into the potential of integrating deep learning, specifically CNNs, into the diagnosis process, resulting in promising outcomes. Our research highlighted the considerable potential of an attention-guided classification approach, which notably improved the Area Under the Receiver Operating Characteristic (AUROC) on a public dataset. Simultaneously, our study showcased the effectiveness of state-of-the-art object detection models in localizing embolisms, further refined by leveraging ensemble techniques. This combination has proven especially effective in detecting smaller embolisms, which are traditionally challenging to identify. The integration of classifier-guided frameworks and ensemble-based models established a benchmark in the detection performance, surpassing previous benchmarks on the FUMPE dataset. Such an integrated approach has not only enhanced the precision of PE detection but also maintained a reliable level of sensitivity, crucial for medical diagnostics. While the outcomes of this research are encouraging, the study also highlighted potential areas of improvement. In summary, this research has established a significant advancement in PE diagnostics through deep learning. As we progress, the broader application of our findings, coupled with continuous refinement based on the identified limitations, holds the potential to significantly improve PE patient care. The fusion of human expertise and AI-driven solutions stands to offer a more holistic, accurate, and efficient diagnosis process, ultimately enhancing patient outcomes and elevating the standards of medical care.

**Funding**

This work was made possible by High Impact grant QUHI-CENG-23/24-216 from Qatar University and is also supported via funding from Prince Sattam Bin Abdulaziz University project number (PSAU/2023/R/1444). The statements made herein are solely the responsibility of the authors. The open-access publication cost is covered by the Qatar National Library.

**CRediT authorship contribution statement**

**Fabiha Bushra:** Conceptualization, Methodology, Data curation, Formal analysis, Software, Writing – original draft, Writing – review & editing. **Muhammad E.H. Chowdhury:** Conceptualization, Methodology, Supervision, Funding acquisition, Writing – original draft, Writing – review & editing. **Rusab Sarmun:** Data curation, Methodology, Writing – original draft, Writing – review & editing. **Saidul Kabir:** Data curation, Methodology, Writing – original draft, Writing – review & editing. **Menatalla Said:** Methodology, Writing – original draft, Writing – review & editing. **Sohaib Bassam Zoghoul:** Validation, Methodology, Writing – original draft, Writing – review & editing. **Adam Mushtak:** Validation, Methodology, Writing – original draft, Writing – review & editing. **Israa Al-Hashimi:** Validation, Writing – original draft, Writing – review & editing. **Abdulrahman Alqahtani:** Conceptualization, Funding acquisition, Writing – original draft, Writing – review & editing. **Anwarul Hasan:** Methodology, Software, Funding acquisition, Writing – original draft, Writing – review & editing.

**Declaration of competing interest**

The authors declare that they have no known competing financial interests or personal relationships that could have appeared to influence the work reported in this paper.

**Data availability**

Data will be made available on request.